\documentclass[11pt]{article}
\usepackage{acl2014}
\usepackage{times}
\usepackage{url}
\usepackage{latexsym}

\usepackage{pdfpages}
\usepackage{graphicx}
\usepackage{caption}
\usepackage{subcaption}
\usepackage{amsmath}
\usepackage{bbm}
\usepackage{color}

\title{Reconstructing Native Language Typology from Foreign Language Usage}

\author{Yevgeni Berzak \\
  CSAIL MIT \\
  {\tt berzak@mit.edu} \\\And
  Roi Reichart \\
  Technion IIT \\
  {\tt roiri@ie.technion.ac.il} \\\And
  Boris Katz \\
  CSAIL MIT \\
  {\tt boris@mit.edu} \\}
\date{}

\begin{document}

\includepdf[pages={1}]{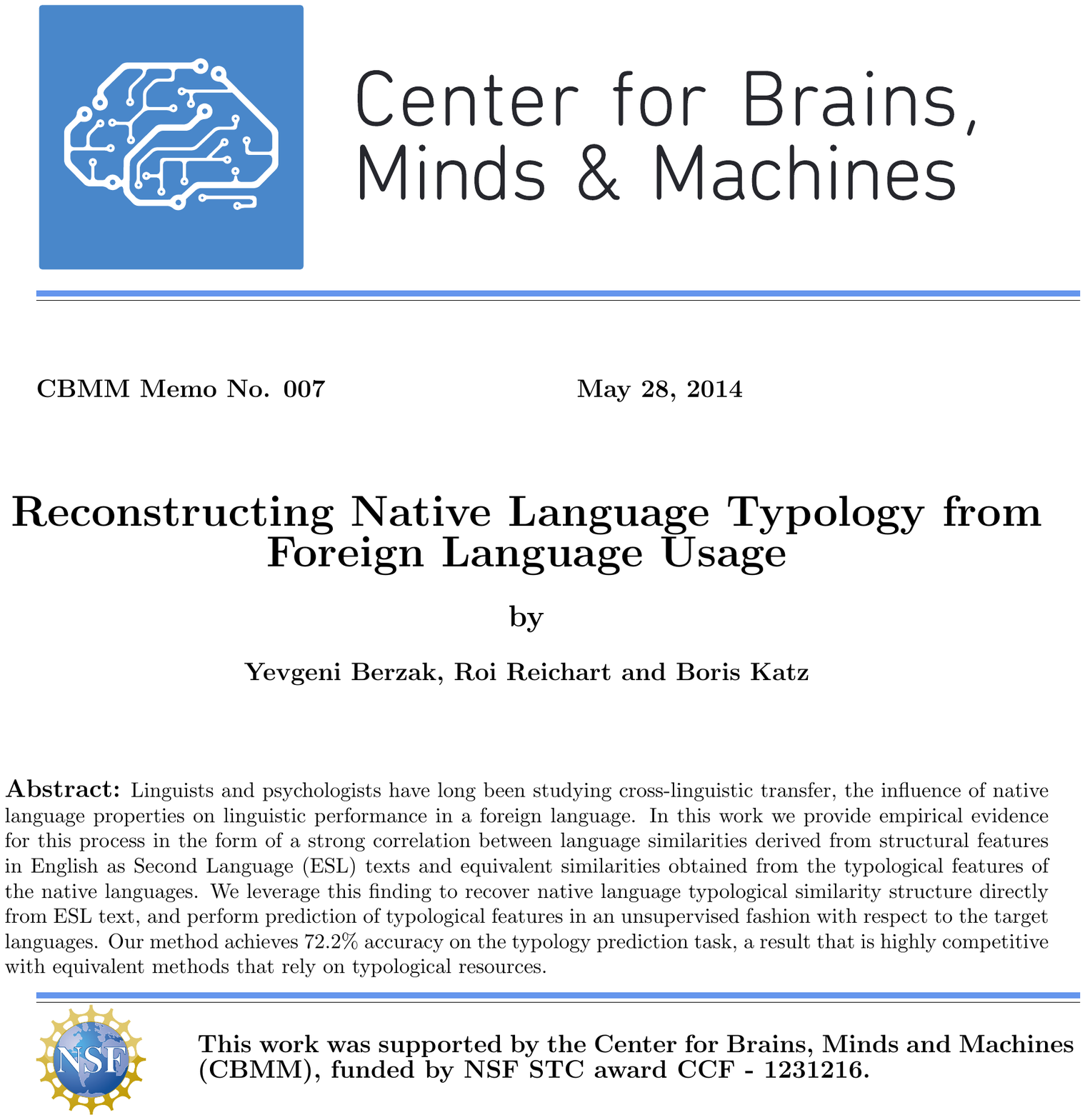}

\maketitle
\begin{abstract}

Linguists and psychologists have long been studying cross-linguistic
transfer, the influence of native language properties on linguistic performance 
in a foreign language. In this work we provide empirical evidence for this process 
in the form of a strong correlation between language similarities derived from 
structural features in English as Second Language (ESL) texts and equivalent 
similarities obtained from the typological features of the native languages. 
We leverage this finding to recover native language typological similarity 
structure directly from ESL text, and perform prediction of typological features 
in an unsupervised fashion with respect to the target languages. 
Our method achieves 72.2\% accuracy on the typology prediction task, 
a result that is highly competitive with equivalent methods that rely on typological 
resources.

\end{abstract}

\section{Introduction}

Cross-linguistic transfer can be broadly described as the application 
of linguistic structure of a speaker's native language in the context of 
a new, foreign language. Transfer effects may be expressed on various levels 
of linguistic performance, including pronunciation, word order, lexical borrowing
and others \cite{jarvis2007transfer}. Such traces are prevalent in non-native English, 
and in some cases are even celebrated in anecdotal hybrid dialect names such 
as ``Frenglish'' and ``Denglish''.

Although cross-linguistic transfer was extensively studied in Psychology, Second 
Language Acquisition (SLA) and Linguistics, the conditions under which it occurs, 
its linguistic characteristics as well as its scope remain largely under 
debate \cite{jarvis2007transfer,gass1992language,odlin1989transfer}. 

In NLP, the topic of linguistic transfer was mainly addressed in relation to the 
Native Language Identification (NLI) task, which requires to predict the native language
of an ESL text's author. The overall high performance on
this classification task is considered to be a central piece of evidence for the 
existence of cross-linguistic transfer \cite{jarvis2012approaching}.
While the success on the NLI task confirms the ability to extract native language 
signal from second language text, it offers little insight into the linguistic 
mechanisms that play a role in this process. 

In this work, we examine the hypothesis that cross-linguistic structure transfer is governed 
by the \emph{typological properties of the native language}. We provide empirical 
evidence for this hypothesis by showing that language similarities derived from 
structural patterns of ESL usage are strongly correlated with similarities obtained 
directly from the typological features of the native languages. 

This correlation has broad implications on the ability to perform 
inference from native language structure to second language performance and vice versa. 
In particular, it paves the way for a novel and powerful framework for \emph{comparing native 
languages through second language performance}. This framework overcomes many of the 
inherent difficulties of direct comparison between languages, 
and the lack of sufficient typological documentation for the vast majority of the 
world's languages.

Further on, we utilize this transfer enabled framework for the task of 
reconstructing typological features. Automated prediction of language typology is 
extremely valuable for both linguistic studies and 
NLP applications which rely on such information \cite{naseem2012parsing,tackstrom2013parsing}. 
Furthermore, this task provides an objective external testbed for the quality of our 
native language similarity estimates derived from ESL texts. 

Treating native language similarities obtained from ESL 
as an approximation for typological similarities, we use them to predict
typological features without relying on typological annotation for the target languages. 
Our ESL based method yields 71.4\% -- 72.2\% accuracy on the typology reconstruction task, 
as compared to 69.1\% -- 74.2\% achieved by typology based methods which depend on 
pre-existing typological resources for the target languages.

To summarize, this paper offers two main contributions.
First, we provide an empirical result that validates the systematic existence of 
linguistic transfer, tying the typological characteristics of the native language
with the structural patterns of foreign language usage.
Secondly, we show that ESL based similarities can be directly used for prediction of 
native language typology. As opposed to previous approaches, our method achieves 
strong results without access to any a-priori knowledge about the target language typology. 

The remainder of the paper is structured as follows. Section \ref{sec:relatedwork} 
surveys the literature and positions our study in relation to previous research on cross-linguistic 
transfer and language typology. Section \ref{sec:data} describes the ESL corpus and the 
database of typological features. In section \ref{sec:eslsim}, we delineate our method for deriving 
native language similarities and hierarchical similarity trees from structural features 
in ESL. In section \ref{sec:walssim} we use typological features to construct another set of language 
similarity estimates and trees, which serve as a benchmark 
for the typological validity of the ESL based similarities. 
Section \ref{subsec:corres} provides a correlation analysis between the ESL based and
typology based similarities. Finally, in section \ref{sec:reconstructionres} we report 
our results on typology reconstruction, a task that also provides an evaluation 
framework for the similarity structures derived in sections \ref{sec:eslsim} and \ref{sec:walssim}.

\section{Related Work}\label{sec:relatedwork}
Our work integrates two areas of research, cross-linguistic 
transfer and linguistic typology. 

\subsection{Cross-linguistic Transfer}
The study of cross-linguistic transfer has thus far evolved in 
two complementary strands, the linguistic \emph{comparative} approach, and the computational 
\emph{detection} based approach. While the comparative approach focuses on case study based qualitative analysis 
of native language influence on second language performance, the detection based 
approach revolves mainly around the NLI task. 

Following the work of Koppel et al. \shortcite{koppel2005nli}, NLI has been gaining increasing interest in NLP, 
culminating in a recent shared task with 29 participating systems \cite{nli2013report}. 
Much of the NLI efforts thus far have been focused on exploring various feature sets for 
optimizing classification performance. While many of these features are linguistically
motivated, some of the discriminative power of these approaches stems from cultural 
and domain artifacts. For example, our preliminary experiments with a typical 
NLI feature set, show that the strongest features for predicting Chinese are strings such 
as \emph{China} and \emph{in China}. Similar features dominate the weights of other languages as well. 
Such content features boost classification performance, but are hardly relevant for modeling 
linguistic phenomena, thus weakening the argument that NLI classification performance is indicative 
of cross-linguistic transfer.

Our work incorporates an NLI component, but departs from the performance optimization
orientation towards leveraging computational analysis for better understanding of the 
relations between native language typology and ESL usage. In particular, our choice of 
NLI features is driven by their relevance to linguistic typology rather than their 
contribution to classification performance. In this sense, our work aims to take a 
first step towards closing the gap between the detection and comparative approaches 
to cross-linguistic transfer.

\subsection{Language Typology}
The second area of research, language typology, deals with the documentation and comparative study of 
language structures \cite{2011typlogyhandbook}. Much of the descriptive work in the 
field is summarized in the World Atlas of Language Structures (WALS)\footnote{\url{http://wals.info/}} \cite{wals}
in the form of structural features. We use the WALS features as our
source of typological information.

Several previous studies have used WALS features for hierarchical clustering of 
languages and typological feature prediction. Most notably, Teh et al. \shortcite{teh2007walsclustering} 
and subsequently Daum{\'e} III \shortcite{daume2009clustering} predicted typological features from language 
trees constructed with a Bayesian hierarchical clustering model. In Georgi et al. \shortcite{georgi2010clustering} 
additional clustering approaches were compared using the same features and evaluation method. 
In addition to the feature prediction task, these studies also evaluated their clustering 
results by comparing them to genetic language clusters. 

Our approach differs from this line of work in several aspects. First, similarly to 
our WALS based baselines, the clustering methods presented in these studies are affected 
by the sparsity of available typological data. Furthermore, these methods rely 
on existing typological documentation for the target languages. Both issues are obviated 
in our English based framework which does not depend on any typological information to 
construct the native language similarity structures, and does not require any knowledge about the target languages 
except from the ESL essays of a sample of their speakers. Finally, we do not compare our 
clustering results to genetic groupings, as to our knowledge, there is no firm theoretical 
ground for expecting typologically based clustering to reproduce language phylogenies. 
The empirical results in Georgi et al. \shortcite{georgi2010clustering}, which show that 
typology based clustering differs substantially from genetic groupings, support this assumption. 

\section{Datasets}\label{sec:data}

\subsection{Cambridge FCE}

We use the Cambridge First Certificate in English (FCE) dataset 
\cite{fcecorpus2011} as our source of ESL data. This corpus is a 
subset of the Cambridge Learner Corpus (CLC)\footnote{\url{http://www.cambridge.org/gb/elt/catalogue/subject/custom/item3646603}}. It contains English 
essays written by upper-intermediate level learners of English for the FCE examination. 

The essay authors represent 16 native languages.
We discarded Dutch and Swedish speakers due to the small number of documents available for
these languages (16 documents in total). The remaining documents are associated with the 
following 14 native languages: Catalan, Chinese, French, German, Greek, Italian, Japanese, 
Korean, Polish, Portuguese, Russian, Spanish, Thai and Turkish. Overall, our corpus 
comprises 1228 documents, corresponding to an average of 87.7 documents per native language. 

\subsection{World Atlas of Language Structures}\label{subsec:wals}

We collect typological information for the FCE native languages from WALS. 
Currently, the database contains information about 2,679 of the world's 
7,105 documented living languages \cite{ethnologue}.
The typological feature list has 188 features, 175 of which are present in our dataset. 
The features are associated with 9 linguistic categories: Phonology, Morphology, Nominal 
Categories, Nominal Syntax, Verbal Categories, Word Order, Simple Clauses, Complex Sentences and Lexicon.
Table \ref{wals-examples} presents several examples for WALS features and their range 
of values.

\begin{table*}[!ht]
\begin{center}
\begin{tabular}{|l|l|l|l|}
\hline
\bf ID & \bf Type           & \bf Feature Name              & \bf Values  \\ \hline
26A    & Morphology         & Prefixing vs. Suffixing in    & Little affixation, Strongly suffixing, Weakly\\
       &                    & Inflectional Morphology       & suffixing, Equal prefixing and suffixing,\\
       &                    &                               & Weakly prefixing, Strong prefixing.\\ \hline
30A    & Nominal            & Number of Genders             & None, Two, Three, Four, Five or more. \\
       &   Categories                 &                               & \\ \hline
83A    & Word Order         & Order of Object and Verb        & OV, VO, No dominant order.  \\ \hline
111A   & Simple Clauses     & Non-periphrastic Causative     & Neither, Morphological but no compound, \\ 
       &                    & Constructions           	    & Compound but no morphological, Both.\\      

\hline
\end{tabular}
\end{center}
\caption{Examples of WALS features. As illustrated in the table examples, WALS features can 
take different types of values and may be challenging to encode.}\label{wals-examples}
\end{table*}

One of the challenging characteristics of WALS is its low coverage, stemming from lack 
of available linguistic documentation. It was previously estimated that 
about 84\% of the language-feature pairs in WALS are unknown \cite{daume2009clustering,georgi2010clustering}. 
Even well studied languages, like the ones used in our work,
are lacking values for many features. For example, only 32 of the WALS 
features have known values for all the 14 languages of the FCE corpus.
Despite the prevalence of this issue, it is important to bear in mind that some 
features do not apply to all languages by definition. For instance, feature 81B 
\emph{Languages with two Dominant Orders of Subject, Object, and Verb} is relevant only to 189 
languages (and has documented values for 67 of them).

We perform basic preprocessing, discarding 5 features that have values for 
only one language. Further on, we omit 19 features belonging to the category Phonology 
as comparable phonological features are challenging to extract from the ESL textual data. 
After this filtering, we remain with 151 features, 114.1 features 
with a known value per language, 10.6 languages with a known value per feature
and 2.5 distinct values per feature.

Following previous work, we binarize all the WALS features, expressing each feature
in terms of $k$ binary features, where $k$ is the number of values the original feature
can take. Note that beyond the well known issues with feature binarization, this strategy is not optimal
for some of the features. For example, the feature 111A \emph{Non-periphrastic Causative 
Constructions} whose possible values are presented in table \ref{wals-examples} would 
have been better encoded with two binary features rather than four. The question of 
optimal encoding for the WALS feature set requires expert analysis 
and will be addressed in future research. 

\section{Inferring Language Similarities from ESL}\label{sec:eslsim}
Our first goal is to derive a notion of similarity between languages 
with respect to their native speakers' distinctive structural usage patterns 
of ESL. A simple way to obtain such similarities is to train a probabilistic 
NLI model on ESL texts, and interpret the uncertainty of this classifier in 
distinguishing between a pair of native languages as a measure of their similarity.

\subsection{NLI Model}
The log-linear NLI model is defined as follows:
\begin{equation}
p(y|x;\theta) = \frac{\exp(\theta \cdot f(x,y))}{\sum_{y' \in Y} \exp(\theta \cdot f(x,y'))}
\end{equation}
where $y$ is the native language, $x$ is the observed English document and 
$\theta$ are the model parameters. 
The parameters are learned by maximizing the L2 regularized log-likelihood of the training data
$D = \{(x_{1}, y_{1}),...,(x_{n}, y_{n})\}$.  
\begin{equation}
L(\theta) = \sum_{i=1}^{n}\log p(y_{i}|x_{i};\theta)-\lambda\lVert\theta\rVert^{2}
\end{equation}
The model is trained using gradient ascent with L-BFGS-B \cite{lbfgs1995}. 
We use 70\% of the FCE data for training and the remaining 30\% for development
and testing. 

As our objective is to relate native language and target language \emph{structures}, 
we seek to control for biases related to the content of the essays. 
As previously mentioned, such biases may arise from the essay prompts as well as 
from various cultural factors. We therefore define the model using only 
\emph{unlexicalized} morpho-syntactic features, which capture structural properties of English usage.

Our feature set, summarized in table \ref{feature-set}, contains features 
which are strongly related to many of the structural features in WALS. In particular,
we use features derived from labeled dependency parses. These features encode properties 
such as the types of dependency relations, ordering and distance between the 
head and the dependent. Additional syntactic information is obtained using POS n-grams. 
Finally, we consider derivational and inflectional morphological affixation.
The annotations required for our syntactic features are obtained from the Stanford POS tagger 
\cite{2003stanfordtagger} and the Stanford parser \cite{2006stanforddep}. The morphological 
features are extracted heuristically. 

\begin{table*}[!ht]
\begin{center}
\begin{tabular}{|l|l|}
\hline
\bf Feature Type                          &\bf Examples          \\ \hline
Unlexicalized labeled dependencies        & Relation = \emph{prep} Head = \emph{VBN} Dependent = \emph{IN} \\
Ordering of head and dependent            & Ordering = \emph{right} Head = \emph{NNS} Dependent = \emph{JJ} \\
Distance between head and dependent       & Distance = $2$ Head = \emph{VBG} Dependent = \emph{PRP} \\
POS sequence between head and dependent   & Relation = \emph{det} POS-between = \emph{JJ} \\           
POS n-grams (up to 4-grams)               & POS bigram = \emph{NN VBZ} \\
Inflectional morphology                   & Suffix = \emph{ing}\\
Derivational morphology                   & Suffix = \emph{ity}\\       
\hline
\end{tabular}
\end{center}
\caption{Examples of syntactic and morphological features of the NLI model. The feature values are set to the
number of occurrences of the feature in the document. The syntactic features 
are derived from the output of the Stanford parser. A comprehensive 
description of the Stanford parser dependency annotation scheme can be found in the Stanford
dependencies manual \cite{2008stanfordmanual}.}\label{feature-set}  
\end{table*}

\subsection{ESL Based Native Language Similarity Estimates}
Given a document $x$ and its author's native language $y$, the conditional probability 
$p(y'|x;\theta)$ can be viewed as a measure of confusion between languages $y$ and $y'$, 
arising from their similarity with respect to the document features. Under this 
interpretation, we derive a language similarity matrix $S'_{ESL}$ whose entries are 
obtained by averaging these conditional probabilities on the training set documents 
with the true label $y$, which we denote as $D_{y}=\{(x_{i}, y) \in D\}$. 
\begin{multline}
S'_{ESL_{y,y'}} = \\
	\begin{cases}
	\frac{1}{\left\vert{D_{y}}\right\vert}\sum\limits_{(x, y) \in D_{y}} p(y'|x;\theta) & \text{if } y' \neq y \\
        1 & \text{ otherwise}
        \end{cases}
\end{multline}

For each pair of languages $y$ and $y'$, the matrix $S'_{ESL}$ contains an entry 
$S'_{ESL_{y,y'}}$ which captures the average probability of mistaking $y$ for 
$y'$, and an entry $S'_{ESL_{y',y}}$, which represents the opposite confusion.  
We average the two confusion scores to receive the 
matrix of pairwise language similarity estimates $S_{ESL}$.
\begin{equation}
S_{ESL_{y,y'}} = S_{ESL_{y',y}} = \frac{1}{2}(S'_{ESL_{y,y'}}+S'_{ESL_{y',y}})
\end{equation}

Note that comparable similarity estimates can be obtained from the 
confusion matrix of the classifier, which records the number of misclassifications
corresponding to each pair of class labels. The advantage of our probabilistic setup
over this method is its robustness with respect to the actual classification performance 
of the model. 

\subsection{Language Similarity Tree}\label{subsec:trees}

A particularly informative way of representing language similarities is in the form 
of hierarchical trees. This representation is easier to inspect than a similarity 
matrix, and as such, it can be more instrumental in supporting linguistic inquiry on 
language relatedness. Additionally, as we show in section \ref{sec:reconstructionres}, 
hierarchical similarity trees can outperform raw similarities when used 
for typology reconstruction.

We perform hierarchical clustering using the Ward algorithm \cite{ward1963}.
Ward is a bottom-up clustering algorithm. Starting with a separate cluster
for each language, it successively merges clusters and returns the tree of cluster
merges. The objective of the Ward algorithm is to minimize the total within-cluster
variance. To this end, at each step it merges the cluster pair that yields the 
minimum increase in the overall within-cluster variance. 
The initial distance matrix required for the clustering algorithm is defined 
as $1 - S_{ESL}$. We use the Scipy implementation\footnote{\url{http://docs.scipy.org/.../scipy.cluster.hierarchy.linkage.html}}
of Ward, in which the distance between a newly formed cluster $a \cup b$ and another 
cluster $c$ is computed with the Lance-Williams distance update formula 
\cite{1967lancewilliams}.

\section{WALS Based Language Similarities}\label{sec:walssim}

In order to determine the extent to which ESL based language similarities 
reflect the typological similarity between the native languages, 
we compare them to similarities obtained directly from the typological features
in WALS. 

The WALS based similarity estimates between languages $y$ and $y'$ are computed 
by measuring the cosine similarity between the binarized typological feature vectors. 
\begin{equation}
S_{WALS_{y,y'}} = \frac{v_{y} \cdot v_{y'}}{\Vert v_{y}\Vert \Vert v_{y'}\Vert }
\end{equation}

As mentioned in section \ref{subsec:wals}, many of the WALS features do not 
have values for all the FCE languages. To address this issue, we experiment 
with two different strategies for choosing the WALS features to be used for 
language similarity computations. The first approach, called \emph{shared-all}, 
takes into account only the 32 features that have known values in all the 14 
languages of our dataset. In the second approach, called \emph{shared-pairwise},
the similarity estimate for a pair of languages is determined based
on the features shared between these two languages.

As in the ESL setup, we use the two matrices of similarity estimates to construct 
WALS based hierarchical similarity trees. Analogously to the ESL case, a WALS based tree 
is generated by the Ward algorithm with the input distance matrix $1 - S_{WALS}$. 
 
\section{Comparison Results}\label{subsec:corres}
 
\begin{figure}
\includegraphics[width=0.47\textwidth]{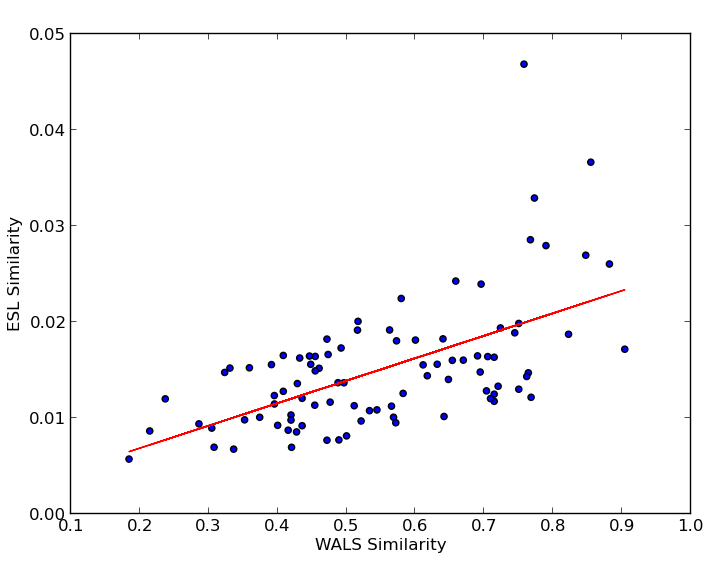}
\caption{\emph{shared-pairwise} WALS based versus ESL based language similarity scores. 
Each point represents a language pair, with the vertical axis 
corresponding to the ESL based similarity and the horizontal axis standing 
for the typological \emph{shared-pairwise} WALS based similarity. The scores correlate 
strongly with a Pearson's coefficient of 0.59 for the \emph{shared-pairwise} 
construction and 0.50 for the \emph{shared-all} feature-set.}
\label{distcor}
\end{figure}

After independently deriving native language similarity matrices 
from ESL texts and from typological features in WALS, we compare them to one another.
Figure \ref{distcor} presents a scatter plot of the language similarities 
obtained using ESL data, against the equivalent WALS based similarities. 
The scores are strongly correlated, with a Pearson Correlation 
Coefficient of 0.59 using the \emph{shared-pairwise} WALS distances and 0.50 using the 
\emph{shared-all} WALS distances.
 
This correlation provides appealing evidence for the hypothesis that 
distinctive structural patterns of English usage arise via cross-linguistic 
transfer, and to a large extent reflect the typological similarities between 
the respective native languages. The practical consequence of this result is 
the ability to use one of these similarity structures to approximate the other. 
Here, we use the ESL based similarities as a proxy for the typological similarities 
between languages, allowing us to reconstruct typological information 
without relying on a-priori knowledge about the target language typology. 

\begin{figure}[!h]
        \begin{subfigure}{0.47\textwidth}
                \includegraphics[width=\textwidth]{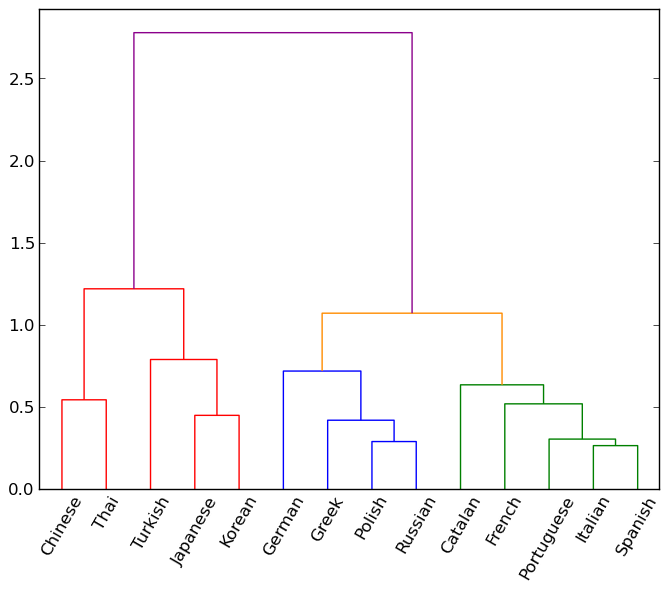}
                \caption{Hierarchical clustering using WALS based \emph{shared-pairwise} distances.}
                \label{walstree}
        \end{subfigure} \\
        \begin{subfigure}{0.47\textwidth}
                \includegraphics[width=\textwidth]{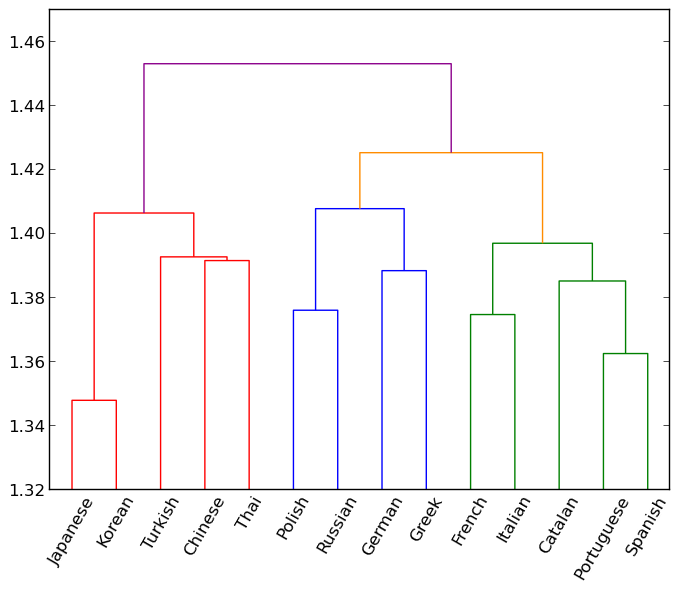}
                \caption{Hierarchical clustering using ESL based distances.}
                \label{engtree}
        \end{subfigure}
        \caption{Language Similarity Trees. Both trees are constructed with the Ward 
		agglomerative hierarchical clustering algorithm. Tree (a) uses the 
		WALS based \emph{shared-pairwise} language distances. Tree (b) uses the ESL derived 
	        distances.}
	\label{simtrees}
\end{figure}

In figure \ref{simtrees} we present, for illustration purposes, the hierarchical 
similarity trees obtained with the Ward algorithm based on WALS and ESL similarities. 
The trees bear strong resemblances to one other. For example, at the top level
of the hierarchy, the Indo-European languages are discerned from the non Indo-European
languages. Further down, within the Indo-European cluster, the Romance languages
are separated from other Indo-European subgroups. Further points of similarity can be
observed at the bottom of the hierarchy, where the pairs Russian and Polish, Japanese and Korean,
and Chinese and Thai merge in both trees.

In the next section we evaluate the quality of these trees, as well as the similarity 
matrices used for constructing them with respect to their ability to support accurate 
nearest neighbors based reconstruction of native language typology.

\section{Typology Prediction}\label{sec:reconstructionres}

Although pairwise language similarities derived from structural features 
in ESL texts are highly correlated with similarities obtained directly 
from native language typology, evaluating the absolute quality of such 
similarity matrices and trees is challenging. 

We therefore turn to typology prediction based evaluation, in which we 
assess the quality of the induced language similarity estimates by their ability to support
accurate prediction of unseen typological features.
In this evaluation mode we project unknown WALS features to a target 
language from the languages that are closest to it in the similarity structure. 
The underlying assumption of this setup is that better similarity structures
will lead to better accuracies in the feature prediction task.


Typological feature prediction not only provides an objective
measure for the quality of the similarity structures, 
but also has an intrinsic value as a stand-alone task. The ability
to infer typological structure automatically can be used 
to create linguistic databases for low-resource languages, 
and is valuable to NLP applications that exploit such resources,
most notably multilingual parsing \cite{naseem2012parsing,tackstrom2013parsing}.

Prediction of typological features for a target language using the language 
similarity matrix is performed by taking a majority vote for the value of each
feature among the $K$ nearest languages of the target language. In case none of the $K$ nearest
languages have a value for a feature, or given a tie
between several values, we iteratively expand the group of nearest languages until neither
of these cases applies.
 
To predict features using a hierarchical cluster tree, we set the value of each target language
feature to its majority value among the members of the parent cluster 
of the target language, excluding the target language itself. For example, using the tree in 
figure \ref{simtrees}(a), the feature values for the target language French will be obtained by 
taking majority votes between Portuguese, Italian and Spanish. Similarly to the matrix based prediction,
missing values and ties are handled by backing-off to a larger set of languages, 
in this case by proceeding to subsequent levels of the cluster hierarchy. For the French 
example in figure \ref{simtrees}(a), the first fall-back option will be the  
Romance cluster.

Following the evaluation setups in Daum{\'e} III \shortcite{daume2009clustering} and Georgi et al. 
\shortcite{georgi2010clustering}, we evaluate the WALS based similarity estimates and 
trees by constructing them using 90\% of the WALS features. We report the average accuracy 
over 100 random folds of the data. In the \emph{shared-all} regime, we provide predictions not only 
for the remaining 10\% of features shared by all languages, but also for all the other
features that have values in the target language and are not used for the tree construction. 

Importantly, as opposed to the WALS based prediction, our ESL based method 
does not require any typological features for inferring language similarities and
constructing the similarity tree. In particular, no typological information is required 
for the target languages. Typological features are needed only for the neighbors 
of the target language, from which the features are projected. 
This difference is a key advantage of our approach over the WALS based
methods, which presuppose substantial typological documentation for all the languages
involved.

Table \ref{completion-table} summarizes the feature reconstruction results.
The ESL approach is highly competitive with the WALS based results, yielding comparable 
accuracies for the \emph{shared-all} prediction, and lagging only 1.7\% -- 3.4\% behind the \emph{shared-pairwise}  
construction. Also note that for both WALS based and ESL based predictions, the highest results
are achieved using the hierarchical tree predictions, confirming the suitability of 
this representation for accurately capturing language similarity structure.

\begin{table}[h]
\begin{center}
\begin{tabular}{|l|l|l|l|}
\hline 
\bf Method                    & \bf NN & \bf 3NN & \bf Tree \\ \hline
WALS \emph{shared-all}        & 71.6   & 71.4    &  69.1 \\
WALS \emph{shared-pairwise}   & 73.1   & 74.1    & \textbf{74.2} \\
ESL                           & 71.4   & 70.7    & \textbf{72.2} \\
\hline
\end{tabular}
\end{center}

\caption{Typology reconstruction results. Three types of predictions are compared,
	nearest neighbor (NN), 3 nearest neighbors (3NN) and nearest tree neighbors (Tree).
	WALS \emph{shared-all} are WALS based predictions, where only the 32 features that have known 
	values in all 14 languages are used for computing language similarities.
	In the WALS \emph{shared-pairwise} predictions the language similarities 
	are computed using the WALS features shared by each language pair. 
	ESL results are obtained by projection of WALS features from the 
	closest languages according to the ESL language similarities.}\label{completion-table}
\end{table}

\begin{figure}[h]
\includegraphics[width=0.47\textwidth]{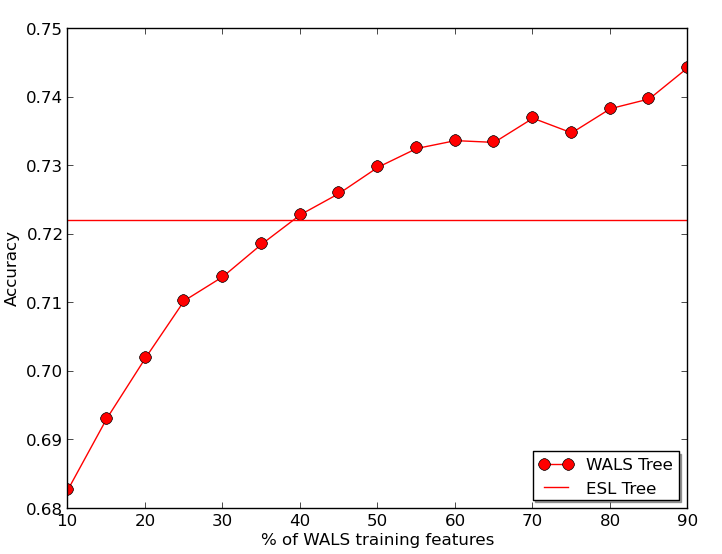}
\caption{Comparison of the typological feature completion performance obtained using the WALS tree with \emph{shared-pairwise} similarities 
and the ESL tree based typological feature completion performance.
	The dotted line represents the WALS based prediction accuracy, while the horizontal line is the ESL based
	accuracy. The horizontal axis corresponds to the percentage of WALS features
	used for constructing the WALS based language similarity estimates.}
\label{completion-graph}
\end{figure}

Figure \ref{completion-graph} presents the performance of the strongest WALS based 
typological feature completion method, WALS \emph{shared-pairwise} tree, as a function of the 
percentage of features used for obtaining the language similarity estimates. 
The figure also presents the strongest result of the ESL method, using the ESL tree, 
which does not require any such typological training data for obtaining the language similarities.
As can be seen, the WALS based approach would require access to almost 40\% of the 
currently documented WALS features to match the performance of the ESL method.

The competitive performance of our ESL method on the typology prediction task underlines its ability to 
extract strong typologically driven signal, while being robust to the partial nature of existing 
typological annotation which hinders the performance of the baselines. Given the small 
amount of ESL data at hand, these results are highly encouraging with regard to the 
prospects of our approach to support typological inference, even in the absence of any 
typological documentation for the target languages. 

\section{Conclusion and Outlook}

We present a novel framework for utilizing cross-linguistic transfer to infer 
language similarities from morpho-syntactic features of ESL text. Trading laborious 
expert annotation of typological features for a modest amount of ESL texts, we are able to 
reproduce language similarities that strongly correlate with the equivalent typology based 
similarities, and perform competitively on a typology reconstruction task. 

Our study leaves multiple questions for future research. For example,
while the current work examines structure transfer, additional investigation
is required to better understand lexical and phonological transfer effects. 

Furthermore, we currently focuse on native language typology, and assume 
English as the foreign language. This limits our ability to study
the constraints imposed on cross-linguistic transfer by the foreign language. 
An intriguing research direction would be to explore other foreign languages and 
compare the outcomes to our results on English.

Finally, we plan to formulate explicit models for the relations between 
specific typological features and ESL usage patterns, and extend our typology induction
mechanisms to support NLP applications in the domain of multilingual processing. 

\section*{Acknowledgments}
We would like to thank Yoong Keok Lee, Jesse Harris and the anonymous reviewers for valuable comments on this paper.
This material is based upon work supported by the Center for Brains, Minds, and Machines (CBMM), funded by NSF STC award CCF-1231216, and 
by Google Faculty Research Award. Roi Reichart was partially supported by the Technion-Microsoft
Electronic-Commerce Research Center.

\bibliographystyle{acl}
\bibliography{acl2014}

\end{document}